\documentclass[twoside,leqno,twocolumn]{article}  
\usepackage{ltexpprt} 

\usepackage{graphicx}

\usepackage{times}
\usepackage{latexsym}
\usepackage{xcolor}
\usepackage{graphicx}
\usepackage{amssymb,amsmath}
\newcommand{\suchthat}{\;\ifnum\currentgrouptype=16 \middle\fi|\;}
\usepackage{lingmacros}
\usepackage{tree-dvips}
\usepackage{amssymb}
\usepackage{amsmath}
\usepackage{graphicx}
\usepackage{float} %pitanje
\usepackage{graphicx}
\usepackage[font=small,skip=0pt]{caption}
\usepackage{wrapfig}
\usepackage{booktabs}
\usepackage{array}
\usepackage{subfigure}
\usepackage{indentfirst}
\usepackage[font={small}]{caption}
\newcolumntype{C}[1]{>{\centering\arraybackslash}p{#1}}

\begin{document}
\title{ \Large Semi-supervised learning for structured regression on partially observed attributed graphs}
\author{Jelena Stojanovic\thanks{Center for Data Anaytics and Biomedical Informatics, Temple University, zoran.obradovic@temple.edu}\\
\and 
Milos Jovanovic\thanks{University of Belgrade, Faculty of Organizational Sciences}\\
\and 
Djordje Gligorijevic\footnotemark[1] \\
\and 
Zoran Obradovic\footnotemark[1]}
\date{}
%\setcounter{chapter}{2} % If you are doing your chapter as chapter one,
%\setcounter{section}{3} % comment these two lines out.
%\title{Structured Regression in Partially Observed Attributed Graphs by Marginalized Gaussian Conditional Random Fields}
%\author{\IEEEauthorblockN{Authors: Triple Blind Review}
%\IEEEauthorblockN{Jelena Stojanovic\IEEEauthorrefmark{1}, Milos Jovanovic\IEEEauthorrefmark{2}, Djordje Gligorijevic\IEEEauthorrefmark{1} and Zoran Obradovic\IEEEauthorrefmark{1}}
%\IEEEauthorblockA{\IEEEauthorrefmark{1}College of Science and Technology, Temple University, Philadelphia, PA, USA
%}
%\IEEEauthorblockA{\IEEEauthorrefmark{2}Faculty of Organizational Sciences, University of Belgrade, Serbia\\
%Email: zoran.obradovic@temple.edu
%}

\maketitle

%\pagenumbering{arabic}
%\setcounter{page}{1}%Leave this line commented out.

\begin{abstract} \small\baselineskip=9pt 
Conditional probabilistic graphical models provide a powerful framework for structured regression in spatio-temporal datasets with complex correlation patterns. However, in real-life applications a large fraction of observations is often missing, which can severely limit the representational power of these models. In this paper we propose a Marginalized Gaussian Conditional Random Fields (m-GCRF) structured regression model for dealing with missing labels in partially observed temporal attributed graphs. This method is aimed at learning with both labeled and unlabeled parts and effectively predicting future values in a graph. The method is even capable of learning from nodes for which the response variable is never observed in history, which poses problems for many state-of-the-art models that can handle missing data. The proposed model is characterized for various missingness mechanisms on 500 synthetic graphs. The benefits of the new method are also demonstrated on a challenging application for predicting precipitation based on partial observations of climate variables in a temporal graph that spans the entire continental US. We also show that the method can be useful for optimizing the costs of data collection in climate applications via active reduction of the number of weather stations to consider. In experiments on these real-world and synthetic datasets we show that the proposed model is consistently more accurate than alternative semi-supervised structured models, as well as models that either use imputation to deal with missing values or simply ignore them altogether. 
\end{abstract}

\section{Introduction}

\begin{figure}[h!]
\centering
\setlength{\abovecaptionskip}{0pt plus 1pt minus 1pt} % Chosen fairly arbitrarily
\setlength{\belowcaptionskip}{0pt plus 1pt minus 1pt} % Chosen fairly arbitrarily
\includegraphics[width=0.45\textwidth]{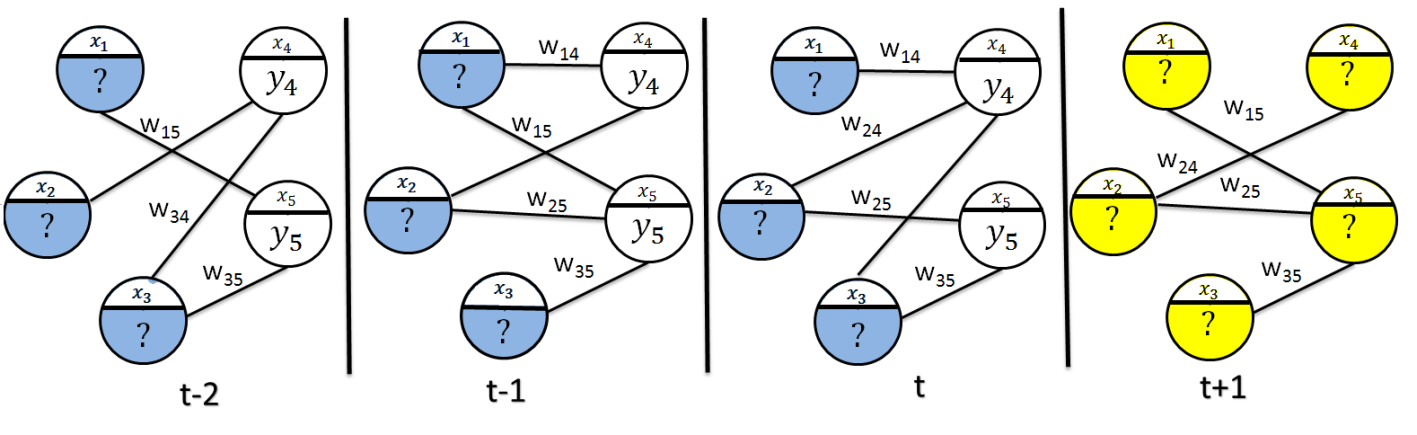}
\vspace{-3pt}
\caption[The LOF caption]{Attributed weighted temporal partially observed graph %\footnotemark
 in which input variables are observed and dependent variables are missing in a large fraction of training data (in blue nodes). The goal is to learn parameters of the model on training data and predict continuous target values of test examples (yellow nodes).}
\vspace{-10pt}
\label{graph}
\end{figure}

Learning and inference with partially observed data is a challenge experienced in many real-world domains. Data is often missing due to sensor failure, reluctance for sharing sensitive information, high cost of collecting the data, or failure of any part of the database. This problem is particularly serious in longitudinal studies when observations on the same units are made repeatedly over time, which is the %, as in such situations the fraction of units with partially missing data is often large. 
situation considered in our article. In particular, as shown in Figure \ref{graph}, we address the problem of structured regression in a temporal graph (prediction of continuous node states in time step $t+1$), where the dependent variable (label) $y$ is missing in a large fraction (up to 80\%) of the training data (time points $1, 2,..., t-1, t$). This constitutes a \textit{semi-supervised learning} (parameter estimation) problem, which is distinct from approaches that try to infer the labels of the unlabeled nodes of a graph \cite{Verbeek, Zhu03}. In our study, an even more challenging problem is considered, where labels at some nodes are missing at all time steps. In addition, each node of a graph is described through a set of explanatory variables $X$ (also called input variables or input attributes), which makes graph \textit{attributed}. The graph is also \textit{temporal} and \textit{weighted} and is observed in discrete snapshots over time, as also exhibited by Figure \ref{graph}.
%Also, note that in Figure \ref{graph} structure is changing over time. 

The nodes in a graphical model are not independent, so ignoring training data with missing labels might disregard too much information. In Figure~\ref{graph}, one can see that if nodes with missing labels are ignored, in this simple example the entire graph structure would be lost and modeling would be limited to unstructured regression or time-series prediction on individual nodes. Utilizing the graph structure may therefore make better use of unlabeled data, especially when lots of nodes have missing labels. In this study, we are considering a discriminative continuous probabilistic graphical model called Gaussian Conditional Random Fields (GCRF) \cite{Radosavljevic2010}. Our goal is to extend the GCRF model to naturally handle missing labels, rather than expecting the missing data to be treated in a preprocessing stage. We propose an extended marginalized GCRF method in which we address the missing label instances by marginalizing out their effect on labeled data, and thus utilizing the information of all observations and preserving the observed graph structure. 

The motivating application that we address in this study is the climate problem of precipitation prediction at the level of individual stations, observed spatially as a graph over time. At some stations measurements of precipitation are missing, but lower resolution predictions of some related climate attributes are provided by climate models. An additional challenge is to estimate future precipitation at additional sites where precipitation is never measured.

The paper is structured as follows. First, in Section~\ref{sec:related_work} we give a brief overview of the existing approaches for handling missing labels.  In Section~\ref{sec:model} we present a marginalized-GCRF model as an extension to the existing GCRF model for handling missing data. The datasets used for evaluation of our method are described in Section~\ref{sec:data}. In Section~\ref{sec:experiments} the experimental setup, factors that influence the performance of the models, the results on synthetic and real-world problems, and their interpretations are presented. Finally, the conclusion is given in Section \ref{sec:conclusions}.
A supplementary file is provided for this paper to elaborate certain topics more in--depth and we refer to it in the following text as \textbf{Appendix}. This file, as well as the m-GCRF code are provided at http://www.dabi.temple.edu/$\sim$zoran/code/sdm15 .

% contains concluding remarks and directions for future work.
\section{Related Work}
\label{sec:related_work}
Treatment of missing data is an old theme in machine learning and statistics literature, and is important because this problem occurs in many real-world datasets. 
Strategies proposed to address this problem are described in the rich literature on this topic \cite{Mcknight2007}. 
In this study we are focused on missing continuous labels in structured regression problems.

One of the standard ways of handling missing values is imputing values based on some predictive model, and then applying the analysis on a fully observed dataset. To exploit the graph structure, previous studies have proposed imputation of missing values based on the exponential random graph model \cite{Ouzienko2011}.  The limitation of such an approach is that it is slow, as it requires Gibbs sampling, and so it cannot handle large graphs. 
Imputation of missing values can also be accomplished using matrix (or tensor) factorization methods. These methods can impute missing values with high accuracy even when large percentages (up to 95\%) of values are missing \cite{Kolda11}. They are also quite fast, allowing the application on dense tensors with a million entries, and sparse tensors with dimensions 1000x1000x1000. However, these methods are not suited for the case when a variable (or node of a graph) is never observed in the dataset, since they cannot recover the factors because there is no enough information \cite{Kolda11}, which is a challenging problem we are considering in this paper. 
Furthermore, imputation-based methods use only point estimates of the missing values, effectively ignoring the prediction uncertainty when learning with imputed values. Techniques known as Multiple Imputation (MI) try to correct for this drawback, by sampling from the posterior distribution of missing values. On the other hand, these techniques can be less effective when a larger fraction of data is missing \cite{Lee2012MI}, and can be computationally very demanding.

Some methods do not require a complete (or imputed) dataset, since they can handle unlabeled data intrinsically. For structured prediction, generative probabilistic models have a natural way for using unlabeled data, since they model the joint distribution of both explanatory and dependent variables. However such approaches have certain drawbacks\cite{Ng01}, which is why discriminative models are often used in practice. On the other hand, with discriminative models it is more difficult to make use of the unlabeled data. Some related studies have approached this problem by creating hybrid discriminative-generative models \cite{Mann2007b}. However, in such hybrid models the number of parameters that need to be estimated is usually large. In addition, these related studies are focused on classification, while our problem of interest is regression \cite{Sokolovska2011}.
Efficient Conditional Random Fields-based methods were also proposed for treating missing labels on graphs \cite{Jiao2006}. However, published methods of that type are applicable only to classification problems \cite{Bellare2009, Jiao2006}. %\cite{Mann2007a,}

In \cite{Zhu03} the authors aim to address the semi-supervised setting that can be used for regression, where the goal was to infer the unknown labels of nodes in a graph, by utilizing a structure derived from the Radial Basis Functions (we compared our approach to this method experimentally).  Another approach that models Gaussian Fields (GF) defined over nearest neighbor graphs in semi-supervised fashion was described in \cite{Verbeek}. In \cite{Verbeek}, authors aim to infer the unknown labels of nodes in a graph, by optimizing parameters using marginal log-likelihood induced from the joint GF density they model. In the experiments, our proposed model is also compared to this model. However, since the authors did not provide the code and our implementation according to the paper \cite{Verbeek} produced poor results, we will not show them in experimental section.

There also exist variants of the conditional graphical CRF models for regression (e.g. the CCRF \cite{Qin2008}, or GCRF \cite{Radosavljevic2010} models). However, these structured regression models are not designed to cope with unlabeled data, other than ignoring the portion of data with missing labels.

\section{The Model}

\label{sec:model}
 \subsection{Gaussian Conditional Random Fields (GCRF)}
\label{sec:gcrf_model}
We are using the class of discriminative models called Gaussian conditional random fields (GCRF) \cite{Radosavljevic2010} for regression in attributed weighted temporal graphs, where explanatory variables $X$ are observed in each node $i$ and a dependent continuous variable $y$ corresponds to the state of the nodes. 
The GCRF models the conditional distribution $P(y|X)$ over all outputs $y$ given all inputs $X$:

\vspace{-5pt}
\begin{footnotesize}
\begin{multline}
\label{eq:gcrf}
P(y| X)= \frac{1}{Z}exp(-\sum_{i=1}^{N} \sum_{k=1}^{K}\
\alpha _{k} (y_i - R_k(X))^2 \\
  - \sum_{l=1}^{L}\sum_{i \sim j} \beta_l {S_{ij}}^{(l)}(y_i - y_j)^{2})
\end{multline}
\end{footnotesize}

\noindent where $\alpha$ and $\beta$ are parameters of the association and the interaction potential, respectively and the normalization term $Z(X,\alpha,\beta)$ is an integral over $y$ of the term in the exponent.

In order to enable efficient training and inference of the GCRF model, association (\ref{eq:ass_pot}) and interaction (\ref{eq:intr_pot}) potentials are modeled as quadratic functions of $y$:

\vspace{-15pt}
\begin{small}\begin{equation} \label{eq:ass_pot}
A(\alpha, y_i , X) = -\sum_{i=1}^{N} \sum_{k=1}^{K}\
\alpha _{k} (y_i - R_k(X))^2
\end{equation}
\vspace{-10pt}
\begin{equation} \label{eq:intr_pot}
I(\beta, y_i , y_j , X) =  - \sum_{l=1}^{L}\sum_{i \sim j} \beta_l {S_{ij}}^{(l)}(y_i - y_j)^{2}
\end{equation}
\end{small}

\vspace{-5pt}
\noindent where $R_k(X)$ are unstructured models (functions that map $X\rightarrow y_i$ for each node in a graph, and are learned as classical regression functions taking only $X$ into account; as a special case, only $X_i$ can be used as $X$) and K is the number of those predictors. The interaction potential is modeled to mark the similarity of two nodes' target values according to user defined measure $S_{ij}^{(l)}$ (that defines the weighted undirected graph structure between labels), where the user is allowed to provide multiple ($L$) similarity measures \cite{Radosavljevic2010}.

This choice of feature functions enables us to represent this distribution as a multivariate Gaussian, which results in the  Gaussian conditional random fields (GCRF) model \cite{Radosavljevic2010}:

\vspace{-13pt}
\begin{small}
\begin{equation}
P(y|X) = \frac{1}{(2\pi)^{\frac{N}{2}}\mid \Sigma\mid^\frac{1}{2}}
exp\left(-\frac{1}{2}(y-\mu)^{T}Q(y-\mu) \right )
\end{equation}
\end{small}
where $Q$ is the inverse covariance (precision) matrix:
\vspace{-3pt}
\begin{small}
\begin{equation}
Q = \left\{\begin{matrix}
2\sum_{k=1}^{K}\alpha_{k} + 2\sum_{h}\sum_{l=1}^{L}\beta_l S_{ih}^{(l)} , i=j
\\
-2\sum_{l=1}^{L}\beta_l S_{ij}^{(l)} , i\neq j
\end{matrix}\right.
\end{equation}
\end{small}

\noindent In our experiments, $Q$ is a block-diagonal precision matrix of $NT$x$NT$ dimension, where $N$ is the number of nodes in the graph and $T$ is the number of time steps over which the graph is observed. This way of building a large $Q$ matrix, consisting of blocks of adjacency matrices corresponding to individual time steps, allows capturing evolving structural changes of the temporal graph (as shown in Figure \ref{graph}, the structure of the graph changes from time step to time step).
 
Since the modeled distribution is multivariate Gaussian, the inference is done by computing the expectations in the matrix form
\begin{small}$\mu = Q^{-1} b$\end{small}
 , where \begin{small}
$
b_i = 2\left(\sum_{k=1}^{K}\alpha_{k} R_{k}(X) \right )
$\end{small}. The learning task is to optimize parameters $\alpha$ and $\beta$ by maximizing the conditional log--likelihood, which is a convex objective, and can be optimized using quasi-newton optimization techniques. To ensure the distribution is Gaussian, the $Q$ matrix must be positive definite. To achieve this, exponential transformation of parameters is used, as suggested in \cite{Qin2008}, to make the optimization unconstrained. The parameters are learned by the gradient based methods, and the partial derivatives of the conditional log--likelihood are \cite{Radosavljevic2010}:

{\color{white}.}
\vspace{-0pt}
\begin{small}
\begin{multline}
\label{eq:partalpha}
\frac{\partial P}{\partial \alpha_{k}}=-\frac{1}{2}(y - \mu)^{T}\frac{\partial Q}{\partial \alpha_{k}}(y - \mu)+\\(\frac{\partial b}{\partial \alpha_{k}}-\mu^{T}\frac{\partial Q}{\partial \alpha_{k}})(y-\mu)+
\frac{1}{2}Tr({Q^{-1}\frac{\partial Q}{\partial \alpha_{k}}})
\end{multline} 
\vspace{-12pt}
\begin{multline}
\label{eq:partbeta}
\frac{\partial P}{\partial \beta_{l}}=-\frac{1}{2}(y + \mu)^{T}\frac{\partial Q}{\partial \beta_{l}}(y - \mu)+
\frac{1}{2}Tr({Q^{-1}\frac{\partial Q}{\partial \beta_{l}}})
\end{multline}
\end{small}

The inference task is straightforward, since GCRF is represented by the multivariate Gaussian distribution. The maximum posterior estimate of $y$ is obtained by computing the expected value $\mu$:
%\vspace{-5pt}
\begin{small}$%\begin{equation}
y_{*} = \underbrace{argmax}_{y}P(y|X)=\mu
$\end{small}.%\end{equation}

\vspace{-5pt}
\subsection{m-GCRF for learning with missing values}
\label{sec:extended_model}
 
Our objective is to utilize the entire observed graph structure in cases with missing labels in data. Ignoring nodes that have missing values with GCRF would mean a loss of information from graph structure and building a conditional distribution on labeled data only. If we decompose the original model based on the availability of the labels, we would have:

\vspace{-8pt}
\begin{small}
\begin{equation}
P\left({\begin{bmatrix}y_L \\ y_U \end{bmatrix}} \suchthat {\begin{bmatrix}X_L\\ X_U\end{bmatrix}}\right) \sim
\mathcal{N}\left(\begin{bmatrix}
\mu_L \\
\mu_U
\end{bmatrix},
\begin{bmatrix}
Q_{LL} & Q_{LU}\\
Q_{UL} & Q_{UU}
\end{bmatrix}^{-1}  \right )
\end{equation}
\end{small}

\noindent where subscript $L$ denotes the labeled part of the dataset, and $U$ the unlabeled. GCRF that ignores missing data (\textbf{i-GCRF}) would therefore have the model based only on labeled data:
%$P(y_{L} | X_{L}) \sim \mathcal{N}(\mu_L , Q_{LL}^{-1}), $

\vspace{-8pt}
\begin{small}
\begin{equation}
P(y_{L} | X_{L}) \sim \mathcal{N}(\mu_L , Q_{LL}^{-1})
\end{equation}
\end{small}
where $Q_{LL}$ is a precision matrix of exclusively labeled data, excluding the influence of unlabeled graph nodes.
 
Instead of ignoring nodes with missing labels, we want to include the information from $x_U$ that is available for those nodes. Marginalization is a challenging task for regression in general graphical models since it requires integration over hidden nodes' values. A standard approach would be to use the EM algorithm which optimizes the lower bound of the likelihood, but since our model is Gaussian, we can use matrix calculations to express the true gradient of the marginal likelihood over the labeled data. EM is also shown not to perform well when a large chunk of information is missing \cite{Salakhutdinov2003}, as it is using only point estimates of the missing labels. Methods for Multiple Imputation (MI) address this problem, but they are computationally demanding as they use sampling to approximate  marginal distributions \cite{Rubin87}, which we can tackle directly in the Gaussian framework.
 
We define a GCRF model that marginalizes over the unlabeled examples as:

\vspace{-9pt}
\begin{small}
\begin{equation}
\label{eq:marg_int}
p(y_L|X_{L},X_{U}) = \int_{y_U}p(y_L, y_U|X_{L}, X_{U})d_u
\end{equation}
\end{small}
%\vspace{-2pt}
As the original distribution is Gaussian, marginalizing over a subset of variables yields another Gaussian distribution \cite{Bishop2006}:
%\vspace{0pt}
\begin{small}
\begin{equation}\label{eq:marg-lik}
p(y_L|X_{L}, X_{U}) \sim \mathcal{N}(\mu_{L}^{*}, Q_{L}^{*-1})
\end{equation}  \end{small}  
%\vspace{-6pt}
with parameters defined as:
%\vspace{-2pt}
\begin{small}
\begin{equation}
\label{eq:marg_params}
\mu_{L}^{*} = \mu_{L}, Q_{L}^{*-1} = \left( Q_{LL} - Q_{LU} Q_{UU}^{-1} Q_{UL} \right )^{-1}
\end{equation}
\end{small}
\noindent The total derivative of the precision matrix is given by:
\begin{small}
\begin{multline}
\label{eq:marginal_derivative}
d Q_{L}^{*} = d Q_{LL} - d Q_{LU} Q_{UU}^{-1} Q_{UL} + \\ Q_{LU} Q_{UU}^{-1} d Q_{UU} Q_{UU}^{-1} Q_{UL} - Q_{LU} Q_{UU}^{-1} d Q_{UL}
\end{multline}
\end{small}
\indent By calculating gradients of \eqref{eq:marg-lik} with respect to the parameters $\alpha$ and $\beta$, we obtain equations \eqref{eq:partalpha} and \eqref{eq:partbeta} with precision matrix defined as $Q_L^{*}$ \eqref{eq:marg_params} and its derivatives calculated as in \eqref{eq:marginal_derivative}, and we can optimize the marginal likelihood over the labeled nodes. This yields a straightforward, but an effective method for using both labeled and unlabeled data, as will be shown in the experimental part of the paper. With this marginalization model (called \textbf{m-GCRF}) information on all links is retained, and the observed attributes of nodes ($x_U$) with missing labels are also included in learning process.
This can be inferred by observing the precision matrix definition \eqref{eq:marg_params} which takes into account the inverse covariance between labeled and unlabeled data $Q_{UL}$ and the covariance of unlabeled nodes $Q_{UU}^{-1}$. Both are calculated with dependency on node attributes ($X$), and carry necessary information on the complete graph structure. The influence that spreads over some highly connected, but unlabeled, nodes is also conserved. 

Moreover, marginalizing takes the whole distribution over the missing values into account and, unlike point estimates, will produce different effects when the uncertainty of the missing variables under the model is high. This can be seen from the equation \eqref{eq:marg_int}, that can be rewritten as:

\vspace{-5pt} 
\begin{footnotesize}
\begin{equation}
\label{eq:marg_int_pros}
p(y_L|X_{L}, X_{U}) = \int_{y_U}p(y_L| y_U,X_{L}, X_{U})p(y_U| X_{L}, X_{U})d_u
\end{equation}
\end{footnotesize}

\noindent The second term under the integral is the modeled distribution of the unlabeled nodes, and can be seen as a prior for the observed likelihood. If the uncertainty of the label estimates for the unlabeled part is very high, this prior acts effectively as a uniform prior and does not affect the distribution over the labeled part. 

\section{Data}
\label{sec:data}
In this section we will describe data-generation process of synthetic graphs and introduce data set from climate domain we will use in Section~\ref{sec:exp_precp_data} to characterize effectiveness of our method and alternative approaches.
\vspace{-7pt}
\subsection{Synthetic data} \label{sec:synth_data}
In total, 494 synthetic datasets were generated to evaluate our proposed model and the benchmark models. Experiments aimed to characterize the accuracy of prediction with missing data for various mechanisms were conducted on a 1600 node graph embedded in a 40x40 grid observed in 5 time steps, where 4 independent time steps were used for training and 1 for testing. In addition, bigger graphs (with 50x50=2500, 70x70=4,900, 100x100=10,000 and 120x120=14,400 nodes) were used to characterize scale up properties of the methods as reported in Appendix~C.

\begin{wrapfigure}{r}[0pt]{0.20\textwidth} %[11]
  \vspace{-15pt}  
  \begin{center}
  \setlength{\abovecaptionskip}{0pt plus 0pt minus 0pt} % Chosen fairly arbitrarily
\setlength{\belowcaptionskip}{0pt plus 0pt minus 0pt} % Chosen fairly arbitrarily
  \includegraphics[width=0.20\textwidth]{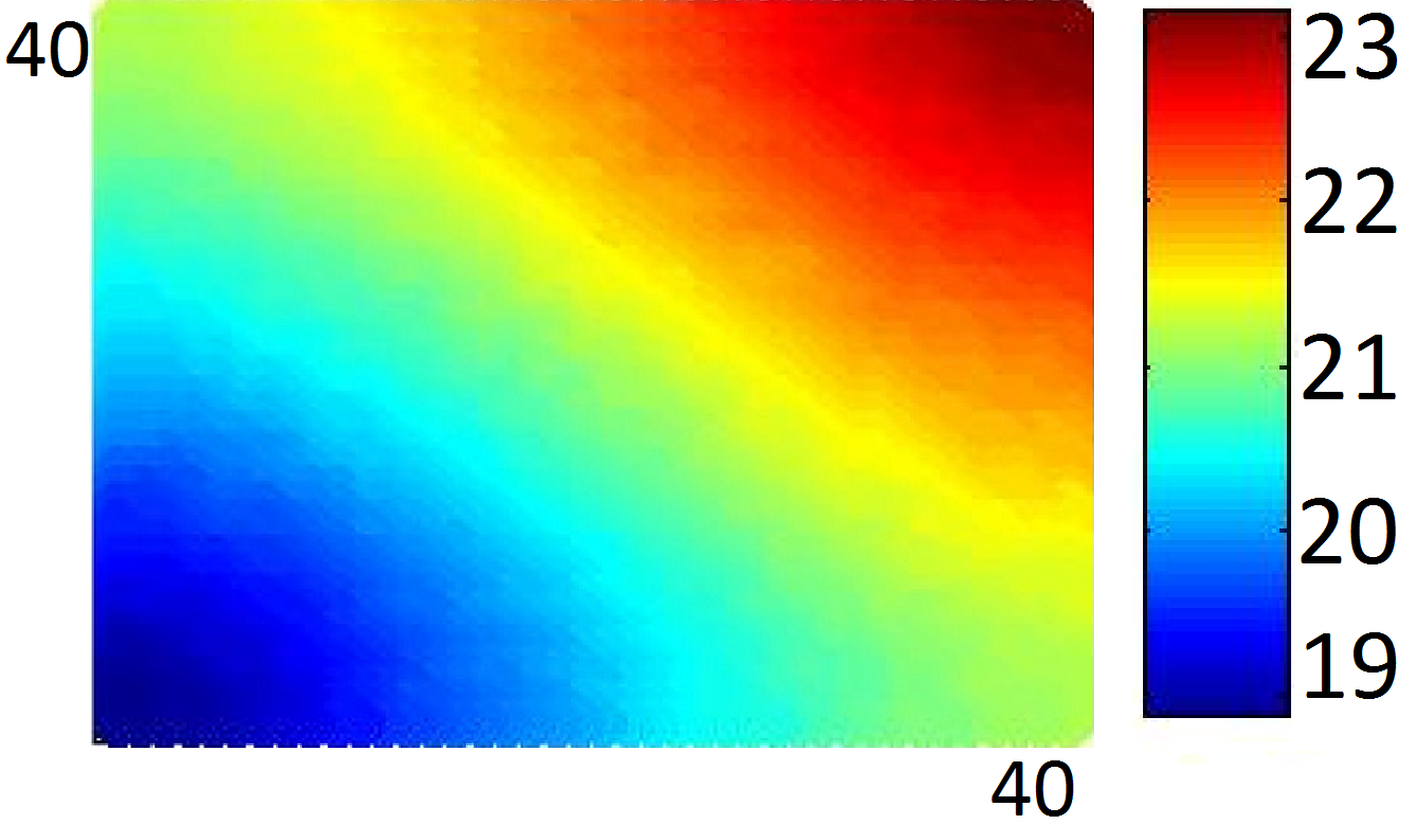}
  \end{center}
  \vspace{-10pt}
  \caption{ Heatmap of the values of response variable $y$ in the grid}
  \vspace{-10pt}
\label{heatmap}
\end{wrapfigure}

Each dataset is constructed using GCRF as a generative process. The unstructured model in this GCRF was a Feed-forward Neural Network (NN) with 30 input variables (in the range 0.01 to 0.1), 60 hidden nodes with sigmoid activation, and a single output. This NN with 10\% additive noise in input is applied to a set of unlabeled examples (30-dimensional tuples) and these examples were distributed on a grid structure based on the value of the NN output, with a tendency for growth of the output values from the lower left to the upper right corner of the grid, as shown at Figure~\ref{heatmap}.

The reason for the described data generation process is an assumption that similar values will be closer positioned in the space (grid). A data similarity matrix containing weights of links between nodes is also generated randomly with weight values in a range from 0.5 to 1. Although GCRF method enables modeling evolving structural changes (as explained in Section \ref{sec:gcrf_model}), in this paper we assume a static structure since the interconnection patterns among nodes in the climate forecasting problem motivating our study is static. This similarity matrix $S$ is used together with the described neural network $R$ to construct a GCRF model \eqref{eq:gcrf}. The dependent variable $y$ is generated by this GCRF model and is used to label all nodes in all datasets (values of $y$ fell in the range from 19 to 23). 

Experiments on the synthetic 40x40 grid data were conducted with 7 different missingness mechanisms:
Random, Weakly connected, Strongly connected, Strongly connected excluding neighbors, Mid-range $y$ values, Remote neighborhood, Extreme $y$ values (detailed descriptions of missingness mechanisms are given in Section \ref{sec:exp_synth}). For each missingness mechanism, seven kinds of data products were constructed with 0 to 80\% of labels missing. In all experiments, a label removed from a node was removed at all training time steps. For each type of data and each fraction of missing labels, 10 such data sets were constructed, to account for the sample variance. Therefore, a total of 490 synthetic datasets were used in the experiments reported in section \ref{sec:exp_synth} (7 types of missingness x 7 fractions of missing values x 10 repeats). 

\vspace{-7pt}
\subsection{Precipitation data} \label{sec:prec_data}
A dataset of precipitation records from meteorological stations across the USA has been acquired from NOAA's National Climate Data Center (NCDC) \cite{Menne09}. 
%The precipitation data includes 1218 meteorological stations. 
Most of these stations are U.S. Cooperative Observing Network stations generally located in rural locations, while some are National Weather Service First-Order stations that are often located in more urbanized environments. 
A temporal graph is constructed such that nodes at each time slice represent 1218 stations. The spatial information is used for calculating similarities (correlations) between stations, but the graph is constructed such that only stations within certain diameter are connected, thus the graphs structure is sparse. 

In addition to precipitation, there are 6 more variables at each node which we use as input attributes for each station. These variables are acquired from the NCEP/NCAR Reanalysis 1 project \cite{Kalnay}, which is using a state-of-the-art analysis/forecast system to predict climate parameters using past data from 1948 to the present (data available on  NOAA website: http://www.esrl.noaa.gov/psd/). These 6 variables are omega (Lagrangian tendency of air pressure), precipitable water, relative humidity, temperature, u-wind, and w-wind (zonal and meridional components of the wind, respectively). Our goal is to make use of these variables and try to exploit inter-dependencies between stations in order to improve the prediction of precipitation amounts in these stations. Since these attributes are obtained on the lower resolution than individual stations we used the values of attributes from the nearest neighbor. To improve predictions, we perform square root transformation of the target variable and did cross-validation during the training to learn the hyper--parameters of the models.

\vspace{-2pt}
\section{Experiments}
\label{sec:experiments}
To evaluate the effectiveness of the m-GCRF model, we are comparing to several benchmarks (detailed description and comparison of alternative methods with which we experimentally compared to is given in the Appendix~B): %\ref{sec:benchmarks}

\textbf{Neural Networks (NN)}
We test the performance of the unstructured predictor (a Neural Network model) which captures the nonlinear influence of input variables $x$, and is effectively ignoring the unlabeled part of the training data. This kind of model is common in the domain of hydrology \cite{Govinderaju10}, including the precipitation prediction domain \cite{Nastos13, Silverman00}. 

\textbf{i-GCRF} We also evaluate the i-GCRF model that utilizes the unstructured predictor (NN) and the available structure over the labeled data as described in Section \ref{sec:extended_model}. 

\textbf{Multiple Imputation (MI)} 
To apply the MI procedure in our experiments we build a predictor (Gaussian process for regression) to infer the missing values based on always available input attributes ($x$ variables). This predictor outputs a predictive distribution (Gaussian) from which we can sample. Five imputed datasets (samples) are then used to train the GCRF model that outputs the final (averaged) structured predictions (The parameters ($\alpha$ and $\beta$) over the samples are then averaged to produce the final model). 
This benchmark method thus utilizes the information from uncertainties in imputed values, and we use it to characterize the importance of knowing these uncertainties. Furthermore, we use MI because it is a sampling method that approximates the direct marginalization (integral \eqref{eq:marg_int_pros}) over the whole distribution of the unknown values \cite{Williams05}, which in many cases is infeasible to compute directly.

\textbf{Gaussian Fields with Harmonic Properties} 
Our method is also compared to a previous semi-supervised and structured model \cite{Zhu03} summarized in Appendix~B.3. %\ref{sec:HGF_cisto}}. 
Since we have unknown labels in both parts of the training data and all of the test data, we tested this approach over all unknown labels. %\eqref{eq:harmonic}, \eqref{eq:w_gaussian} 
We calculated weight matrix as described in Appendix~(2.4), and used this weight matrix and labeled examples to infer values of unlabeled examples, as described in Appendix (2.1), consisting of test nodes and training nodes with missing labels. Then we can measure the performance of this method on test nodes in order to compare with alternatives. This approach produced poor results on our datasets ($R^2$ up to 0.35 for all missingness mechanisms) and therefore we are not showing these results on figures together with the rest of alternative methods. 

\textbf{Gaussian Fields with Harmonic Properties as an imputation method (HGF-GCRF)}
To better utilize Gaussian Fields with Harmonic Properties, we used it to infer the values of the missing labels in the training data only. Inference about unlabeled data is done using labeled examples and a defined graph structure in each time step. Then we used GCRF on this imputed data to utilize both the input features and the known structure in order to produce predictions on the test data (as described in Appendix~B.2). This approach was named HGF-GCRF in our experiments.%\ref{sec:HGCRF} 

First we evaluated the described methods using synthetic data. For each experiment, we generated synthetic graphs of a certain type, each emphasizing the impact of some data properties on the effect of models we compare, as will be described. All experiments using synthetic graphs are repeated on 10 instances of a graph type in order to analyze the variance of the results. Finally, we validate the effectiveness of the methods on a real-world climate application for precipitation prediction, where the missing labels are present in the observed graph history that we use for training. In both types of datasets, nodes of a graph are completely unlabeled in history, which makes the task more challenging.

The results are shown in terms of mean and standard deviation of $R^2$ as the accuracy measure (1 is the best result and 0 is the mean prediction; shown on y--axes of the following figures) for 0 to 80\% of missingness (on x--axes of the following figures) in data for the proposed m-GCRF model and previously mentioned benchmark models: i-GCRF, HGF-GCRF, Multiple Imputation, as well as an unstructured Neural Network model.
 
\subsection{Characterization on 494 Synthetic Spatial Graphs} \label{sec:exp_synth}
\begin{figure}[h!]
\centering
\setlength{\abovecaptionskip}{5pt plus 1pt minus 1pt} % Chosen fairly arbitrarily
\setlength{\belowcaptionskip}{0pt plus 1pt minus 1pt} % Chosen fairly arbitrarily
\includegraphics[width=0.45\textwidth]{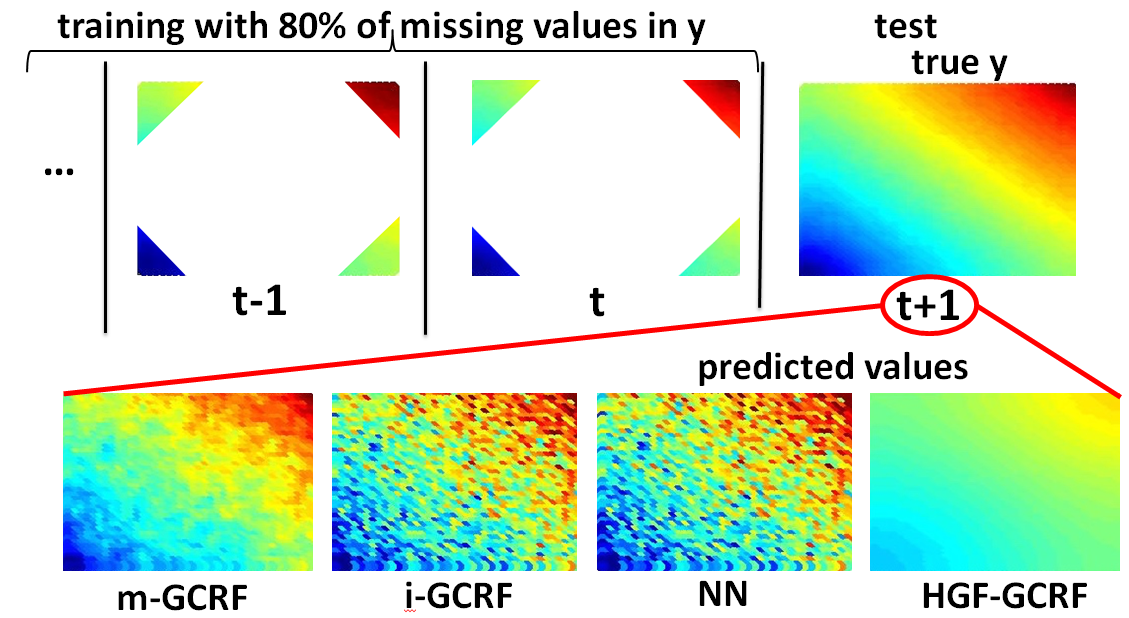}
\vspace{-3pt}
\caption{Predictions (second row in the figure) of the models trained on the data with 80\% of missing values}
\vspace{-10pt}
\label{slicica}
\end{figure}

Prediction results (time step $t+1$) of the models trained ($1, 2... , t$) on data with 80\% of missing values for one of the missingness mechanisms (Experiment 5, Figure~\ref{missing_centar_synth}), are shown in Figure~\ref{slicica} as an example. It is clear from the figure that m-GCRF is able to reconstruct the values in the best way comparing to the other models (results for MI procedure are not shown since this model had negative $R^2$). In the following sections we are going to describe experiments for all missingness processes, but for the lack of the space, figures similar to Figure~\ref{slicica} are omitted and the results are shown in terms of mean and standard deviation of $R^2$ for different fractions of missing data.
% and not in terms of predicted values for each fraction and each mechanism.
\vspace{-5pt}

\subsubsection{Node labels missing at random}
In \textbf{Experiment 1} the objective was to examine how the models will perform in the case when labels are missing completely at random, i.e. where there is no control over the missingness process of nodes. For this experiment 10 40x40 grid-based graphs observed over 5 time steps are used, as explained in Section~\ref{sec:synth_data}. 

\begin{figure}[h!]
\centering
\setlength{\abovecaptionskip}{5pt plus 1pt minus 1pt} % Chosen fairly arbitrarily
\setlength{\belowcaptionskip}{0pt plus 1pt minus 1pt} % Chosen fairly arbitrarily
\includegraphics[width=0.35\textwidth]{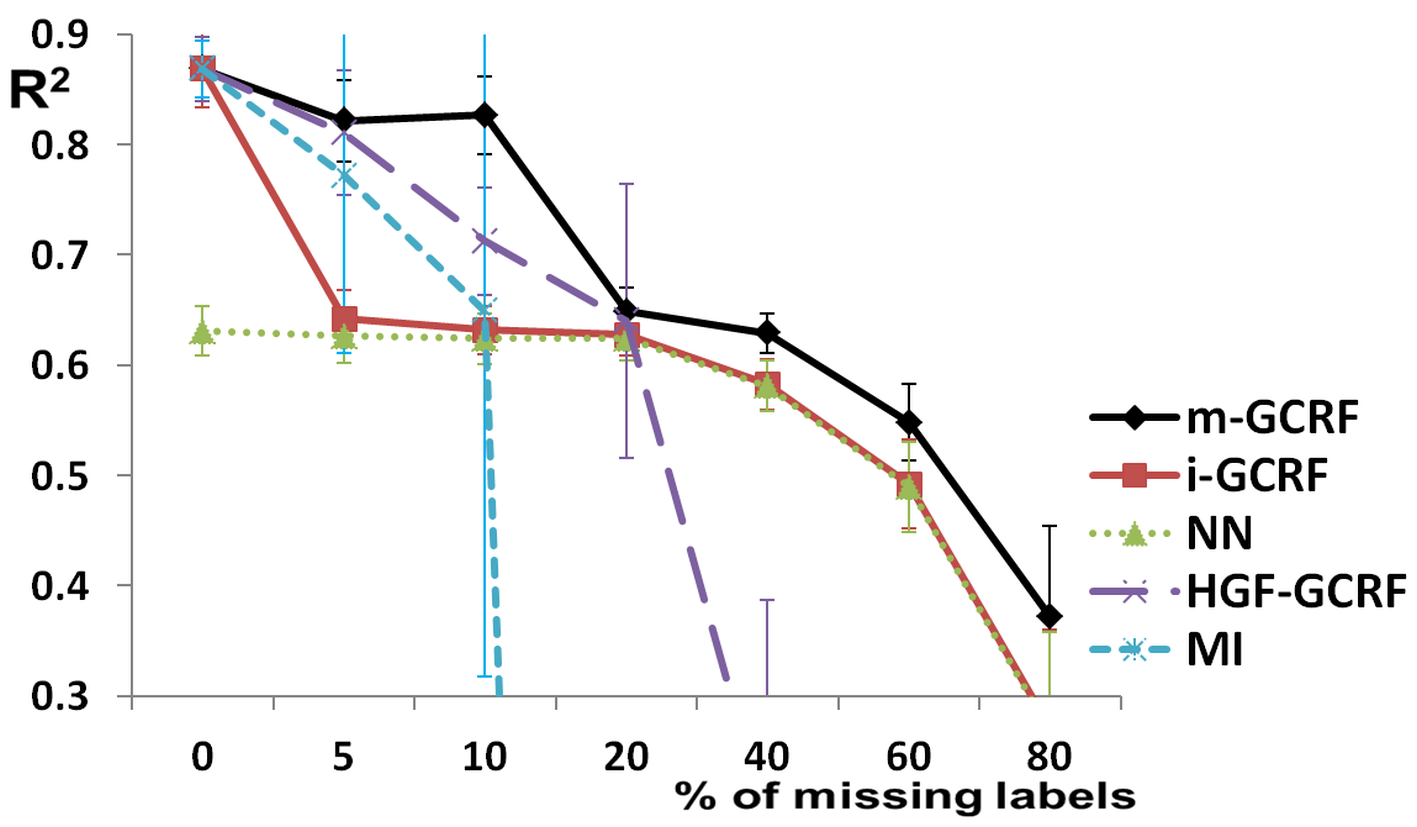}
\vspace{-3pt}
\caption{Accuracy ($R^2$) of the five models when labels are missing completely at random}
\vspace{-10pt}
\label{mcar_synth}
\end{figure}
From the results of Experiment 1 (Figure~\ref{mcar_synth}) we first see that under fully labeled data (0\% missing), both i-GCRF and the extended m-GCRF model, performed significantly better (more than 20\% larger $R^2$) than the unstructured Neural Network model, showing that the grid structure carries a significant amount of information about the label values. By increasing the percentage of missing data, we find that m-GCRF was consistently more accurate than other considered methods (i-GCRF, HGF-GCRF, MI, and NN). In this scenario, the strategy of ignoring unlabeled data is losing information from the structure after only 10\% of missing data, whereas the marginalization approach seems to be more resilient to missing labels, up to 20\%. Using any data imputation method was better than ignoring information about the unlabeled part of the dataset when a small fraction of labels was missing. However, these approaches failed when there was more than 10\% (for MI) or 20\% (for HGF-GCRF) of missing data. Also, we found that imputation-based methods were not stable, since the standard deviation of $R^2$ for these models was large.
\vspace{-5pt}
\subsubsection{Missing labels of weakly connected nodes in a graph}

The goal of \textbf{Experiment 2} was to determine the effect of removing less structurally important nodes. We started by removing the least connected nodes, i.e. the nodes whose total sum of weights is minimal (smallest weighted node degree). 
%For example, this may correspond to a situation when spatially isolated meteorological stations where shut down because of maintenance costs. 
We are assuming that these nodes will not greatly compromise the structure of the graph.

\begin{figure}[h!]
\centering
\setlength{\abovecaptionskip}{5pt plus 1pt minus 1pt} % Chosen fairly arbitrarily
\setlength{\belowcaptionskip}{0pt plus 1pt minus 1pt} % Chosen fairly arbitrarily
\includegraphics[width=0.35\textwidth]{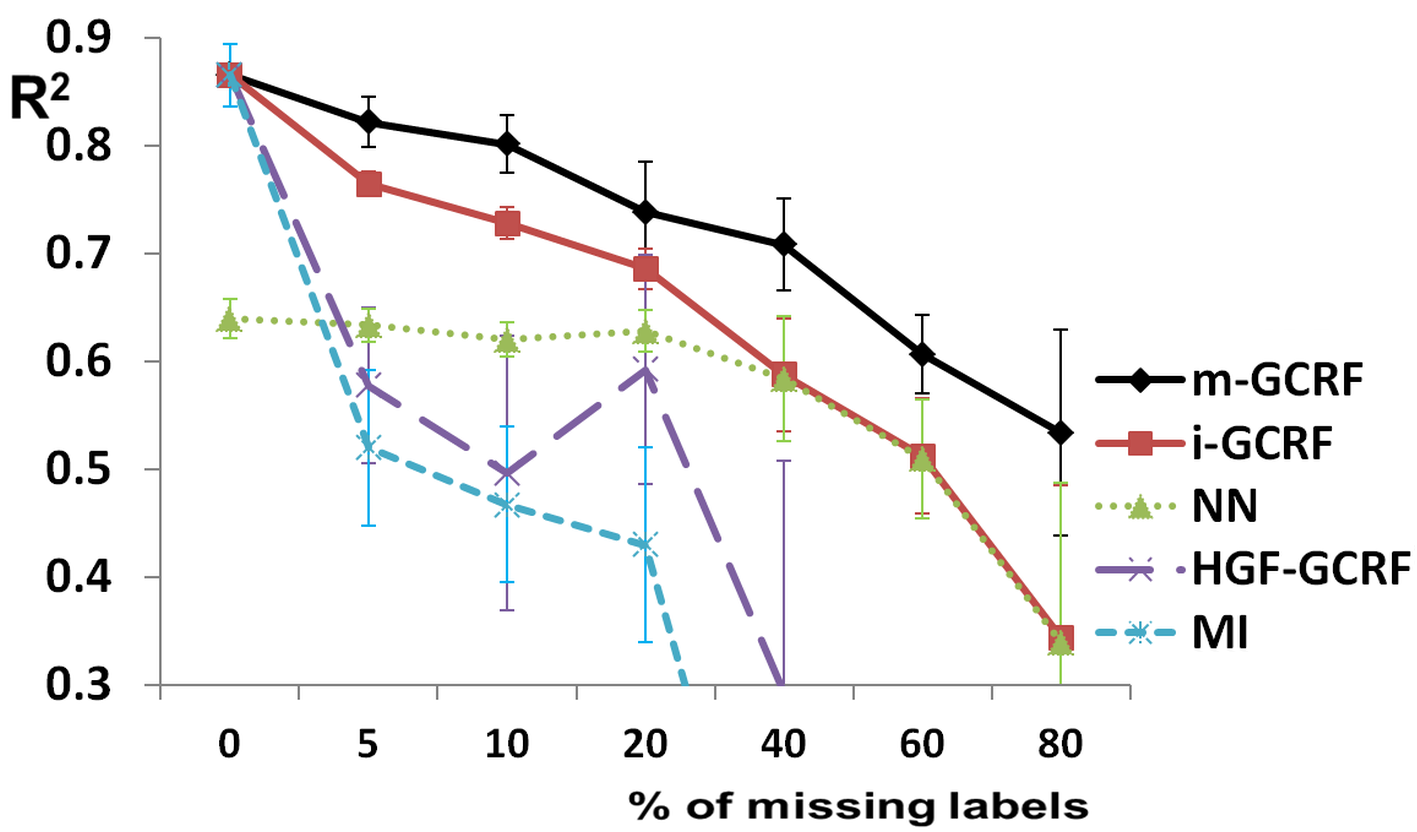}
\vspace{-3pt}
\caption{ Accuracy ($R^2$) of the five models when less connected nodes (structurally less important) are missing }
\vspace{-10pt}
\label{nebitni1_synth}
\end{figure}

Here, in contrast to Experiment 1, we found (Figure~\ref{nebitni1_synth}) that i-GCRF was more accurate than NN for all fractions of missing data. That is to be expected, since the effect of removing first weakly connected nodes is that the remaining structure was more informative, as compared to removing nodes completely at random. Additionally, in these experiments m-GCRF retained good accuracy even when a large percentage of (less connected) nodes was missing, greatly outperforming the i-GCRF method. We found that removing weakly connected nodes hurts HGF-GCRF's accuracy, especially when there is a small percentage of missing labels, since imputation with Gaussian fields will smooth values of less connected nodes too much. We observed that this method becomes more accurate when excluding nodes that are more connected to their neighbors, but since the fraction of labeled data is not large, the HGF-GCRF method is not able to reconstruct all values correctly using only point estimate predictions. We found out that the MI method is a poor choice here, and that variance of such estimates was large. We also note that although the variance of different models seems to overlap, in each instance of the 10 experimental trials the ranking of the models was the same.
\vspace{-5pt}
\subsubsection{Missing labels of strongly connected nodes in a graph}

\begin{figure}[h!]
\centering
\setlength{\abovecaptionskip}{5pt plus 1pt minus 1pt} % Chosen fairly arbitrarily
\setlength{\belowcaptionskip}{0pt plus 1pt minus 1pt} % Chosen fairly arbitrarily
\includegraphics[width=0.35\textwidth]{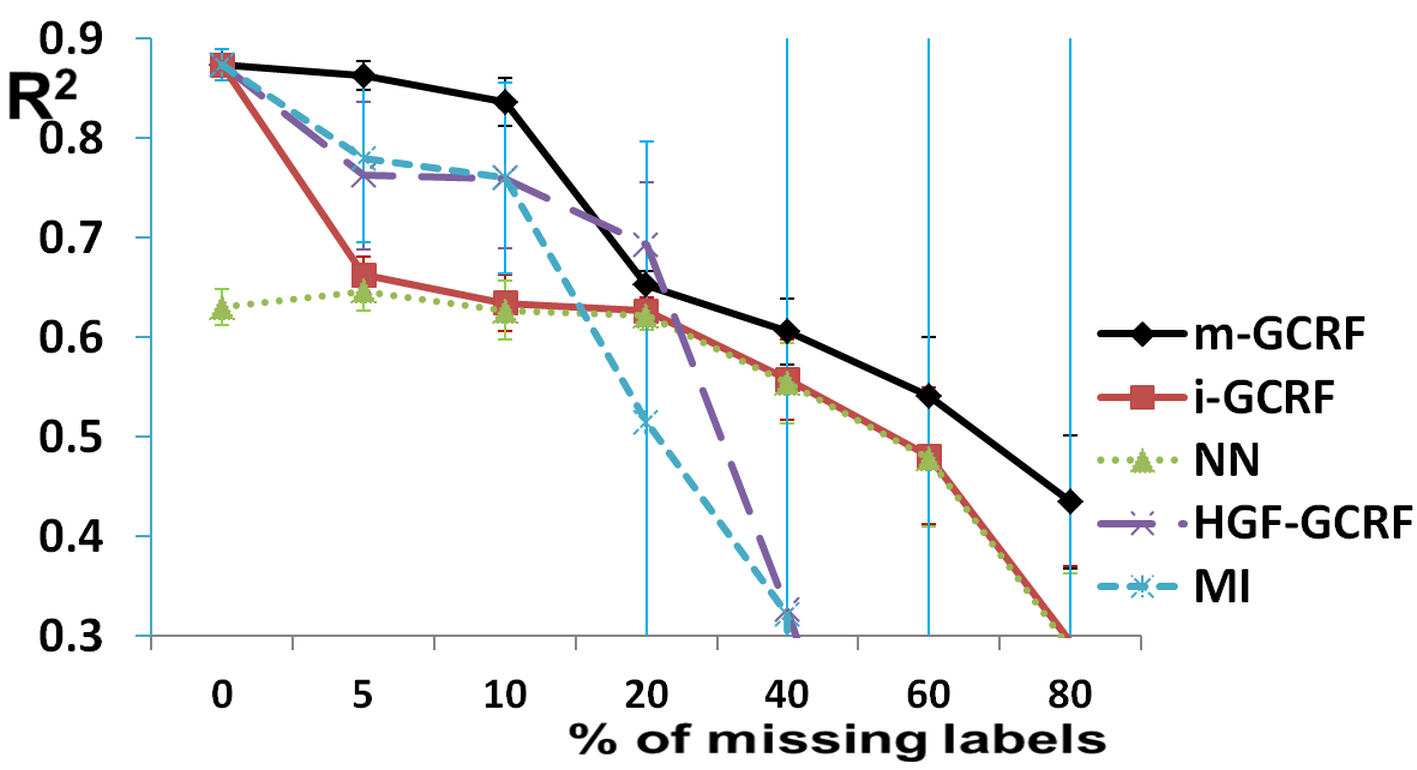}
\vspace{-3pt}
\caption{ Accuracy ($R^2$) of the models when strongly connected nodes are missing }
\vspace{-10pt}
\label{bitni1_synth}
\end{figure}

In \textbf{Experiment 3}, models were evaluated for the case when nodes that are strongly connected (larger weighted node degree) with their neighbors are missing (Figure~\ref{bitni1_synth}). This is the opposite scenario from Experiment 2, and so methods aimed to recover values of missing labels based on structure should be more accurate in such applications.

In Figure~\ref{bitni1_synth} we see a more significant difference (20\% of $R^2$) between i-GCRF and m-GCRF even for graphs with 5\% missing data. This shows that ignoring nodes with missing values that are structurally important is a bad strategy, even for small percentages of missing data. 
As expected in this scenario, HGF-GCRF was able to capture dependencies between these strongly connected nodes and use these connections to rebuild the missing values. 

Another interesting scenario is examined in \textbf{Experiment 4}, where nodes that are missing are strongly connected (as in Experiment 3), but we never removed the neighboring nodes, so the Markov blanket of each node is preserved.
\begin{figure}[h!]
\centering
\setlength{\abovecaptionskip}{5pt plus 1pt minus 1pt} % Chosen fairly arbitrarily
\setlength{\belowcaptionskip}{0pt plus 1pt minus 1pt} % Chosen fairly arbitrarily
\includegraphics[width=0.35\textwidth]{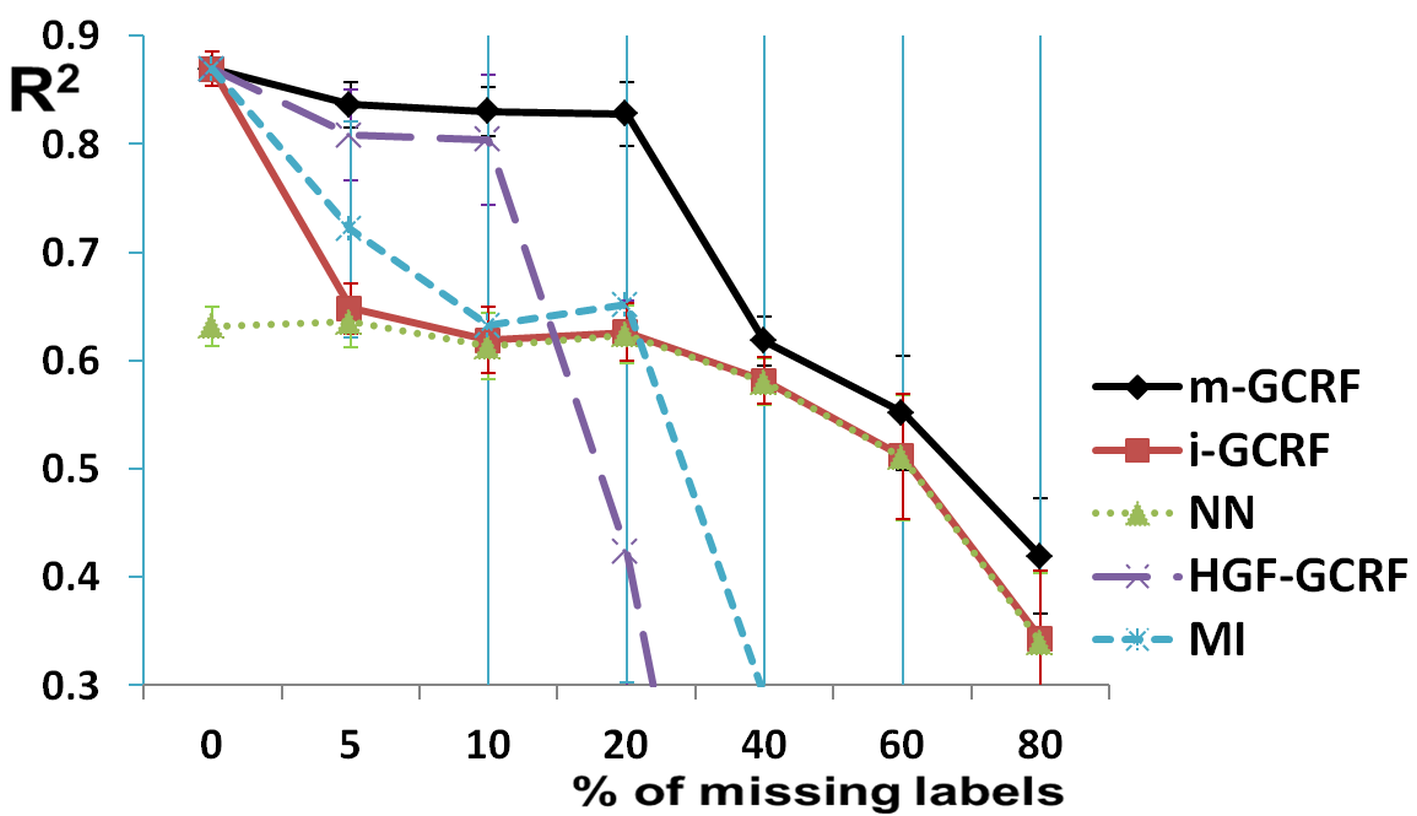}
\vspace{-2pt}
\caption{ Accuracy ($R^2$) of the models when strongly connected, but not neighboring, nodes are missing }
\vspace{-10pt}
\label{bitniMarkov_synth}
\end{figure}

The results from this scenario (Figure~\ref{bitniMarkov_synth})  imply that if the neighborhood of each missing node is known, the node could be recovered with more certainty, and m-GCRF can sustain better accuracy on larger percentages of missing labels. This means that if we can influence the mechanism of missingness (e.g. we need to choose how to reduce the labeled training set), this aspect should not be neglected. We found a similar pattern when imputing data using GF (inference about unlabeled nodes via smoothing of labeled neighborhood in this situation really makes sense). However, since it is using only information from structure and point estimation, the method accuracy was lower when more than 10\% of labels were missing and structure was compromised. 

\vspace{-5pt}
\subsubsection{Missing labels of entire neighborhoods}
\begin{figure}[h!]
\centering
\setlength{\abovecaptionskip}{5pt plus 1pt minus 1pt} % Chosen fairly arbitrarily
\setlength{\belowcaptionskip}{0pt plus 1pt minus 1pt} % Chosen fairly arbitrarily
\includegraphics[width=0.35\textwidth]{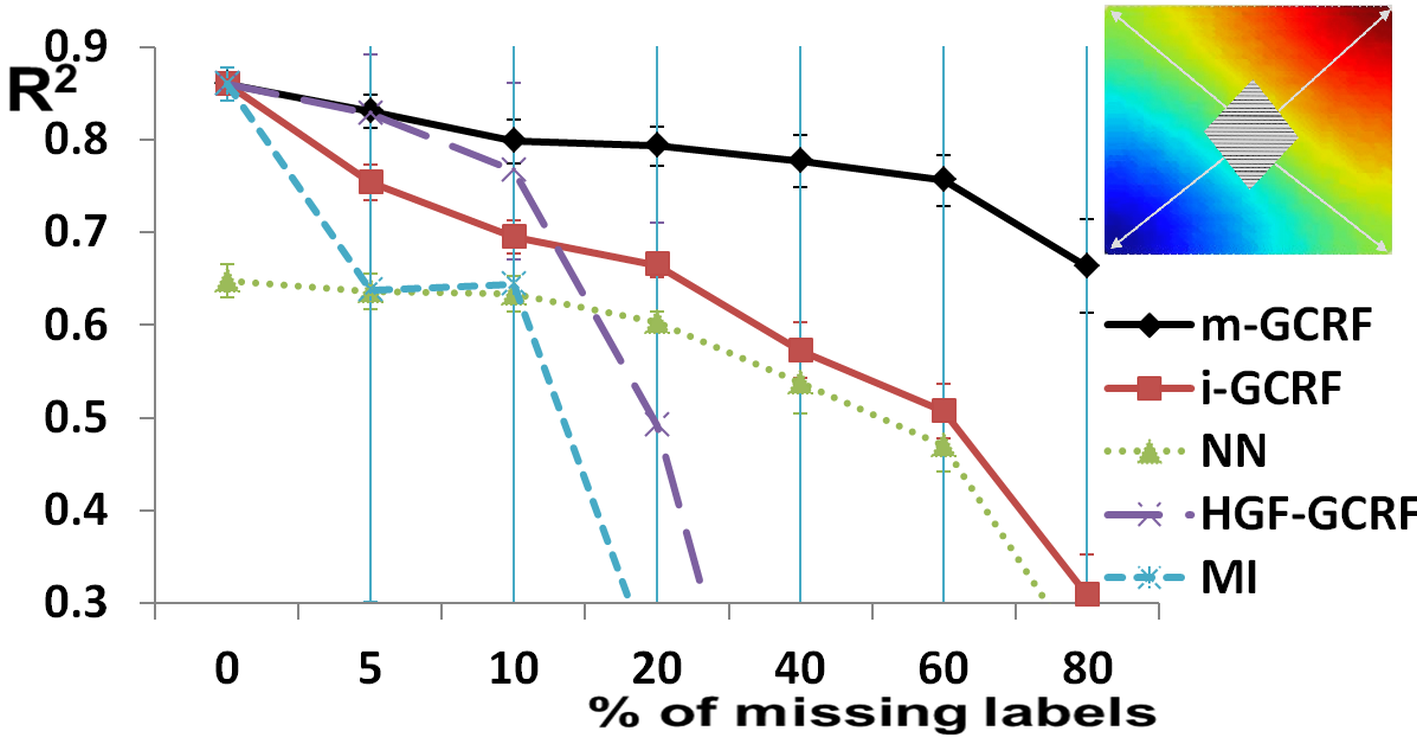}
\vspace{-2pt}
\caption{ Missing labels from center of the grid, where all the extreme values get preserved even for high levels of missingness}
\vspace{-10pt}
\label{missing_centar_synth}
\end{figure}

\begin{figure}[h!]
\centering
\setlength{\abovecaptionskip}{5pt plus 1pt minus 1pt} % Chosen fairly arbitrarily
\setlength{\belowcaptionskip}{0pt plus 1pt minus 1pt} % Chosen fairly arbitrarily
\includegraphics[width=0.35\textwidth]{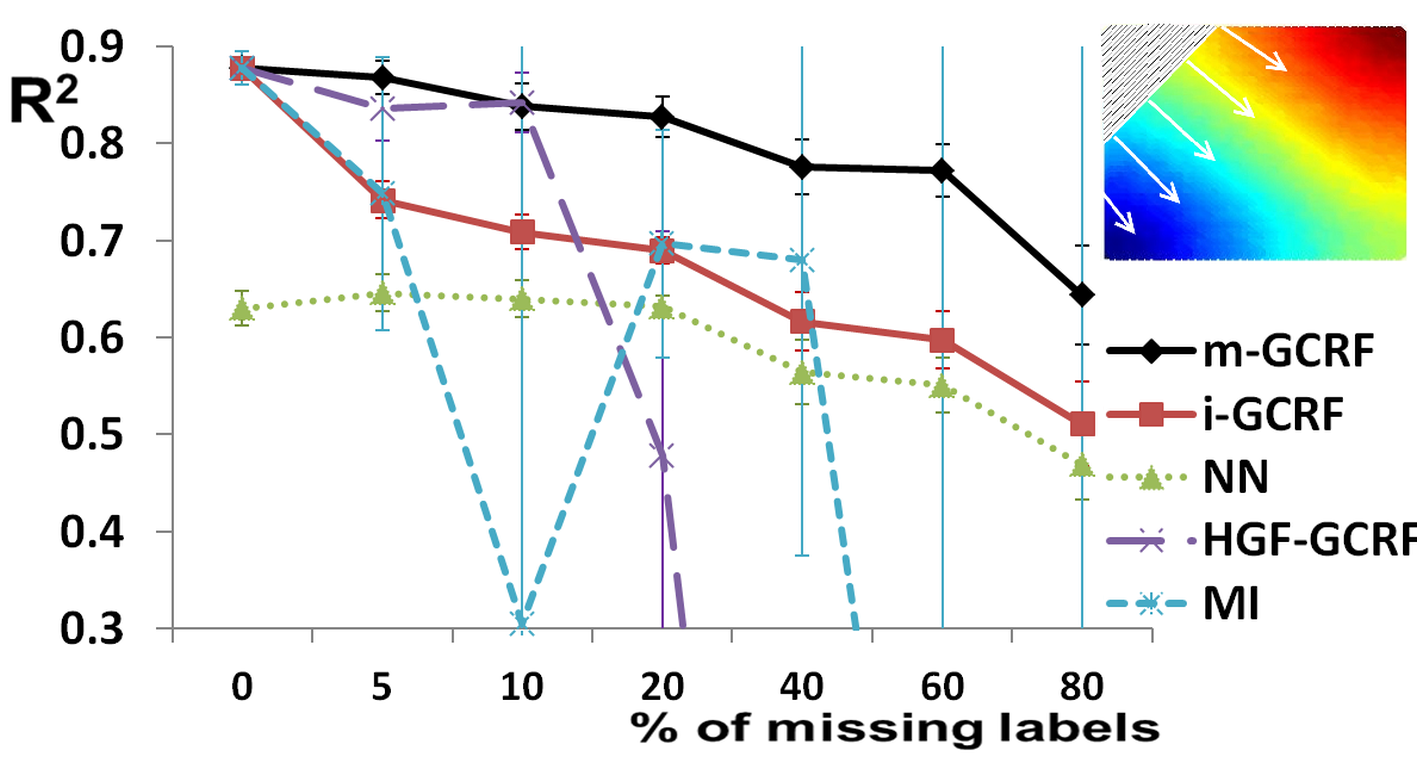}
\vspace{-3pt}
\caption{Missing labels from upper left corner, where there are mostly mid-valued nodes, but soon spreading to a whole grid }
\vspace{-10pt}
\label{missing_from_top_left_synth}
\end{figure}

\begin{figure}[h!]
\centering
\setlength{\abovecaptionskip}{5pt plus 1pt minus 1pt} 
\setlength{\belowcaptionskip}{0pt plus 1pt minus 1pt} 
\includegraphics[width=0.35\textwidth]{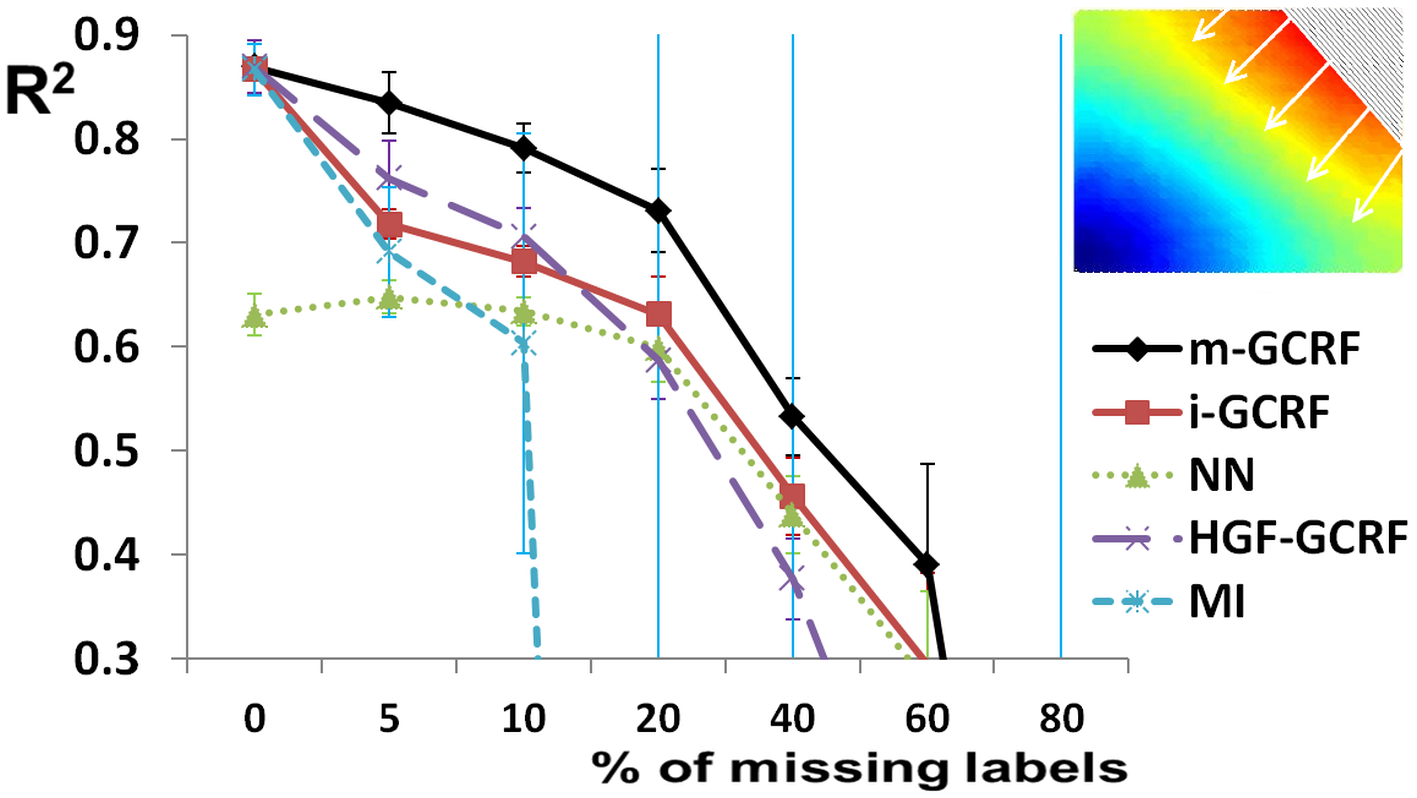}
\vspace{-3pt}
\caption{Labels missing from the part of the grid where only large values accumulate, introducing bias to the models }
\vspace{-10pt}
\label{missing_ekstremi}
\end{figure}

\textbf{Experiments 5, 6 and 7} are aimed to evaluate algorithms when the cause of missingness is in the neighborhood structure. For example, when sensors start going down in a chain reaction from a particular sensor, which, for instance, is caused by spreading fire. In particular, Experiment 5 evaluates algorithms when data starts missing from the center of the grid structure and expands further out (Figure~\ref{missing_centar_synth}). Experiment 6 evaluates accuracy when missingness starts from the upper left corner of the grid structure, where there are mostly middle-range values of the response variable y (Figure~\ref{missing_from_top_left_synth}). Experiment 7 is characterizing a situation when data starts missing from the upper right corner of the grid (Figure~\ref{missing_ekstremi}), where values of the response variable y are largest.

In all three situations, we notice improvements in the accuracy from the unstructured predictor in both i-GCRF and m-GCRF, since the nodes that are missing are missed along with their neighboring nodes. Therefore, the rest of the structure is fairly well preserved and even i-GCRF should benefit from information of the existing structure. We also see good performance of HGF-GCRF when small chunks of data are missing, and very unstable performance of the MI method. In Experiment 5, since the data starts missing from the center of the grid, we only remove mid-ranged values, and the extreme values of the response variable $y$ are preserved. In the results (Figure~\ref{missing_centar_synth}) we see a huge difference between m-GCRF and i-GCRF on 40-80\% of missing data. In Experiment 6, data of one kind (middle-range values) is omitted first, but after 40\% missing data the extremes (high and low values of $y$) also start missing (they are not preserved as in Experiment 5). Therefore, we see a drop in performance of m-GCRF when more than 60\% of data is missing, such that missingness affects both the highest and lowest values. Finally, Figure~\ref{missing_ekstremi} from Experiment 7 shows that if nodes are missing mainly with large values (extremes) of $y$, this will cause high bias in the estimators, and the performance will drop significantly. This corresponds to the situation of data Missing Not At Random, and is known to have this effect theoretically \cite{Mohan2013}.
\vspace{-2pt}
\subsection{Climate Application: Precipitation Prediction}
\label{sec:exp_precp_data}  
In precipitation data described in Section~\ref{sec:prec_data}, there are no missing values in input variables, but about 5\% of the dependent variables (precipitations) are missing. Our experiments on this data (Figure~\ref{rainDistribution} for the fraction of missing values labeled ''Natural''--(about 5\%)) provide evidence that structured models bring some accuracy improvement versus using an unstructured NN model. Consequently, the graph structure (spatial similarity) carries useful information that structured models were able to exploit. When comparing structured models, we also found that using m-GCRF was beneficial, as additional useful information is extracted by marginalizing missing labels instead of ignoring such cases.
%with $X$ inputs and $Y$ hidden nodes

In follow-up experiments with precipitation data (together with those in Appendix~D) we explored two scenarios inspired by real-world situations in which there would be more missing labels. In one of these scenarios the fraction of missing values naturally increases, while in the other we are asked to reduce data collection in a way that minimizes the information loss. 
\vspace{-7pt}
\subsubsection{Naturally increased missing labels}
\begin{figure}[h!]
\vspace{-5pt}
\centering
\setlength{\abovecaptionskip}{5pt plus 1pt minus 1pt} % Chosen fairly arbitrarily
\setlength{\belowcaptionskip}{0pt plus 1pt minus 1pt} % Chosen fairly arbitrarily
\includegraphics[width=0.35\textwidth]{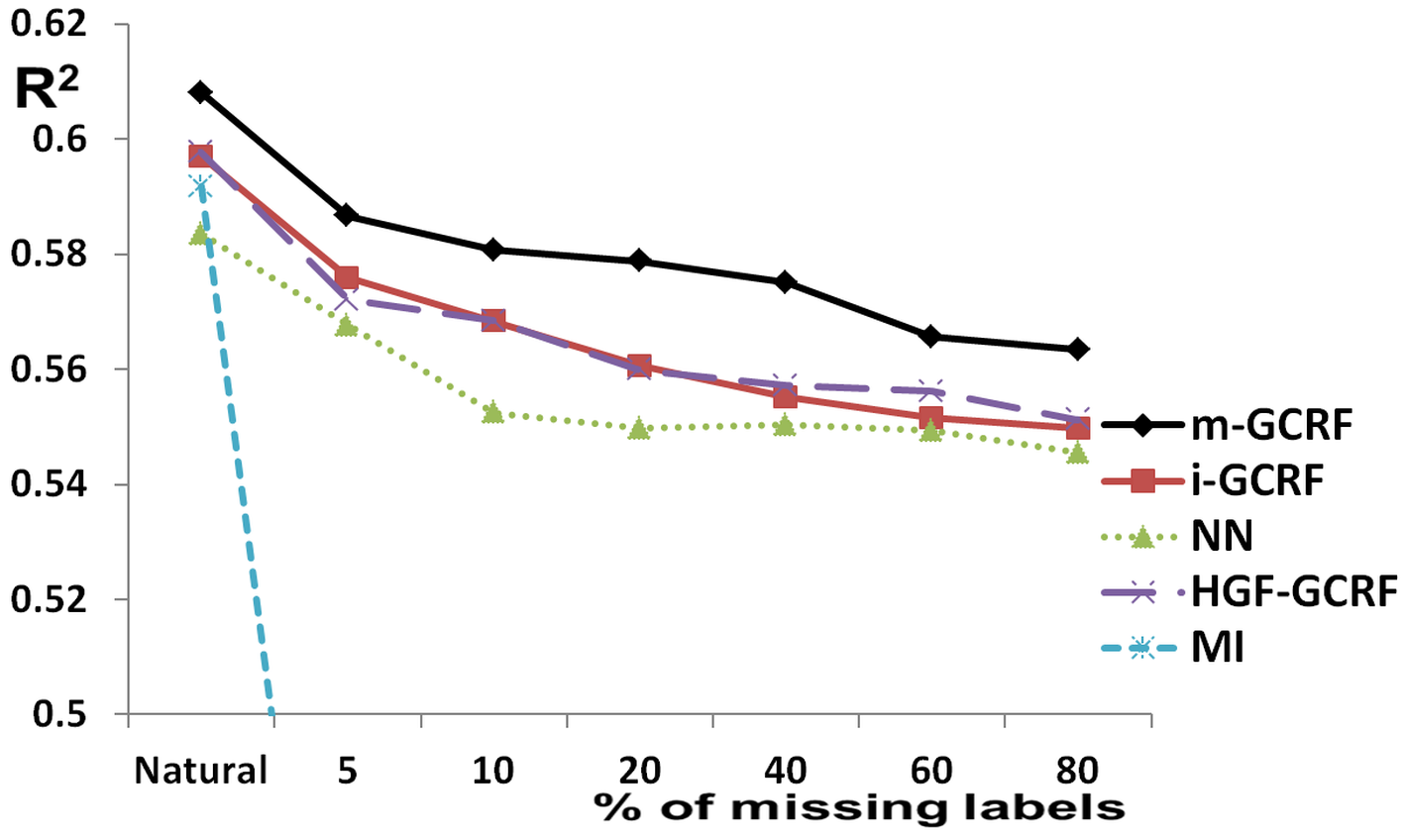}
\vspace{-3pt}
\caption{Accuracy ($R^2$) of all models on precipitation dataset with additional missing labels according to the ''natural'' missingness process}
\vspace{-10pt}
\label{rainDistribution}
\end{figure}

In Precipitation \textbf{Experiment 8} the objective was to examine how these five models would perform if we observed even more missing data. We modeled the probability distribution of originally missing nodes and used it to randomly add more missing values on nodes that are more likely to lose labels according to this distribution. So, this experiment explores the scenario of increasing missing data that could naturally occur under a process similar to the original data missingness mechanism. The results of this experiment are shown in Figure~\ref{rainDistribution}, where labels on the $x$ axis correspond to the fraction of additional missing labels. Here we again see very similar behavior as in the synthetic data experiments described in Section~\ref{sec:exp_synth}. 
By increasing the percentage of missing data, we find that m-GCRF was consistently more accurate than other considered methods (i-GCRF, HGF--GCRF, MI, and NN). i-GCRF and the extended m-GCRF model, performed better than the NN model, showing that the spatial similarity carries a significant amount of information about the label values.
Imputation--based methods were also better than unstructured model when a small fraction of labels was missing.
However, the MI approach failed when there was more than 10\% of missing labels. Using imputation with HGF--GCRF is a marginally better option than ignoring approach, but is not as good as using m-GCRF.
Additional experiments were conducted with different missingness mechanisms on this data with all these methods, but because of limited space, the results are shown in Appendix~D.
%raindistribution
\vspace{-7pt}
\subsubsection{Active restriction of labels}
Finally, in \textbf{Experiment 9} we explore the scenario in which the objective is to reduce the total number of labels in the dataset for future data collection. A practical situation of this kind arises when there is a need to reduce the cost of collecting the meteorological data by closing some stations or learning with spatial interpolation on non-existing stations on a lower scale. By examining how models behave under different control of missingness mechanisms, we can significantly help decision-making regarding the  relevance of various weather stations for accuracy of the overall predictive model. 

\begin{figure}[h!]
\centering
\setlength{\abovecaptionskip}{5pt plus 1pt minus 1pt} % Chosen fairly arbitrarily
\setlength{\belowcaptionskip}{0pt plus 1pt minus 1pt} % Chosen fairly arbitrarily
\includegraphics[width=0.35\textwidth]{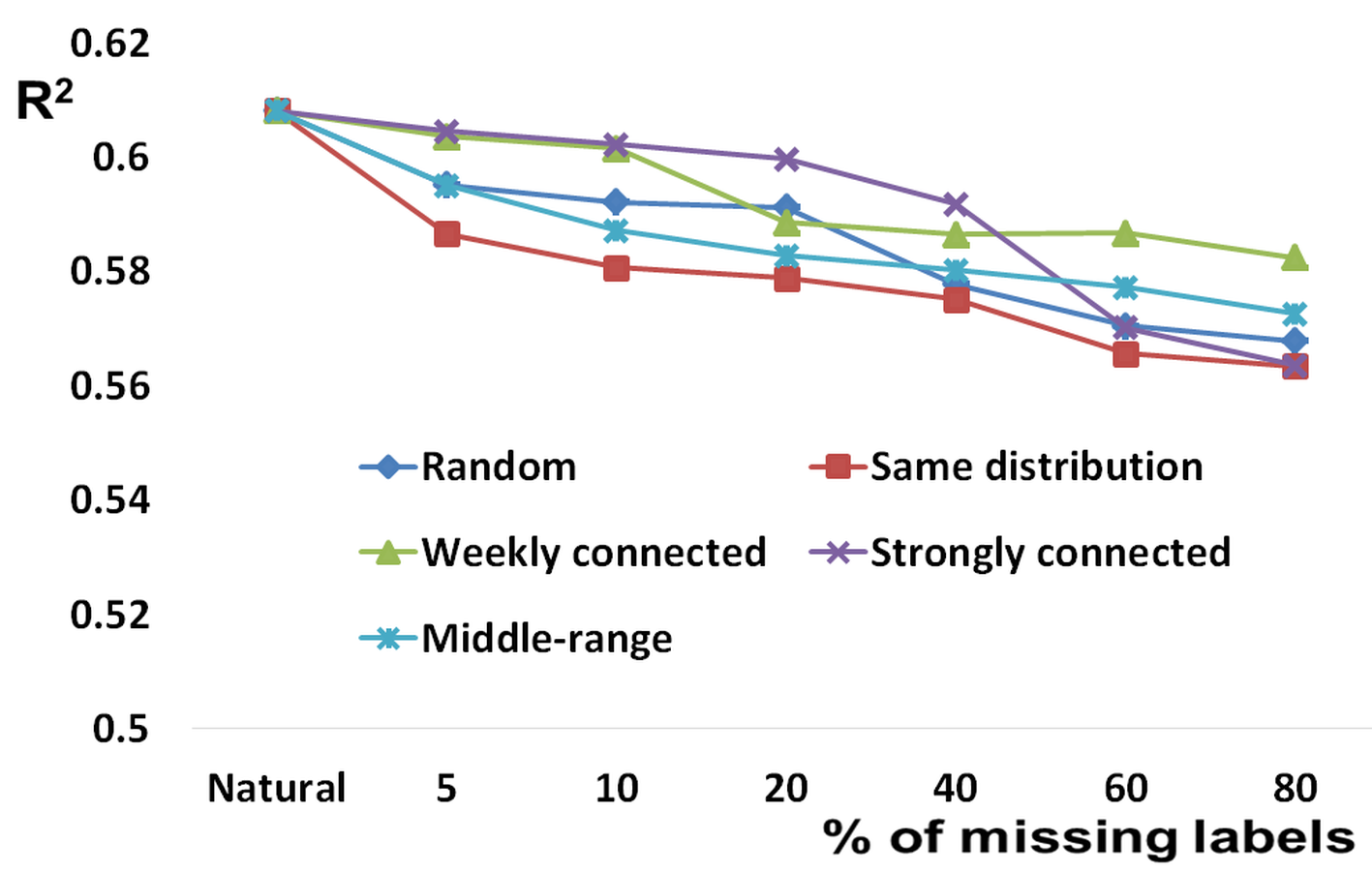}
\vspace{-3pt}
\caption{Accuracy of m-GCRF under different strategies of actively removing labels with additional missing labels}\captionsetup{belowskip=0pt}
\vspace{-10pt}
\label{rain_all}
\end{figure}

We considered several missingness mechanisms. First, we removed nodes at random, as in Experiment 1. Also, we removed weakly connected, or conversely, strongly connected nodes, as in Experiments 2 and 4. Since the connections are determined spatially, this means that strongly connected nodes are ones where there are more stations in the vicinity. Here we also used the strategy of removing labels from nodes that are not neighboring, thus preserving the Markov blanket of each node. Finally, we explored the strategy of removing labels from nodes that historically did not have extreme precipitation values, as in Experiment 5. The results are shown in Figure~\ref{rain_all}. Note that the results in this figure are shown only for m-GCRF, since the objective was to determine which data reduction mechanism results in the largest accuracy of m-GCRF prediction (but the comparison of all five methods is given in Appendix~D).

We found that to control prediction error due to data missingness, removing a large fraction of nodes at random or according to the natural missingness distribution is a bad choice, since it affects the accuracy the most. Instead, it is better to remove strongly connected nodes (without removing their neighboring nodes) if removing less than 40\% of data. We can interpret this as discarding precipitation measuring stations that have many other nearby stations, but keeping the neighbors, since the strongly correlated neighbors are useful in reconstruction of the missing values at the removed stations.

%\vspace{-5pt}
\section{Conclusion}
\label{sec:conclusions}
Longitudinally collected structured data often has a large fraction of missing values. Moreover, nodes of a graph might be completely unlabeled in the history, which makes the task more challenging. For regression in such situations, we propose a m-GCRF model. Our experiments on about 500 spatio-temporal graphs with up to 80\% of missing values provide evidence that m-GCRF is consistently more accurate under various missingness mechanisms than an alternative i-GCRF model that ignores unlabeled data, and than in the domain commonly used unstructured nonlinear regression model. Experiments also show that the proposed model outperformed alternative imputation-based methods. The m-GCRF model is successfully applied to a challenging problem of predicting precipitation based on a temporal graph with missing observations. We also show that if there is a need to actively decrease the amount of labels in the data (e.g. because of the cost of labeling), certain data reduction strategies can be more effective, as they introduce less error when using m-GCRF for prediction.

\vspace{-5pt}
\section*{Acknowledgments} This research was supported by DARPA Grant FA9550-12-1-0406 negotiated by AFOSR, National Science Foundation through major research instrumentation grant number CNS-09-58854.


\vspace{-5pt}
\begin{thebibliography}{99}
\vspace{-10pt}


\bibitem{Kolda11}
Acar E., Dunlavy D. M., Kolda T. G., Mørup M., 'Scalable Tensor Factorizations for Incomplete Data', Chemometrics and Intelligent Laboratory Systems (2011).

%\bibitem{Kolda10}
%Acar E., Dunlavy D. M., Kolda T. G., Mørup M., 'Scalable tensor factorizations with missing data', in: {\em Proc. of SDM2010}, SIAM, pp. 701–-712, (2010).

\bibitem{Bellare2009}
Bellare,K. and McCallum, A., 'Learning extractors from unlabeled text using relevant databases', in {\em Proc. of the Sixth International Workshop on IIWeb}, (2007).

\bibitem{Bishop2006}
Bishop, C. M. {\em Pattern Recognition and Machine Learning}, Springer-Verlag New York, (2006).

\bibitem{Govinderaju10}
Govindaraju R.S., Rao A.R., {\em Artificial Neural Networks in Hydrology}, Springer Publishing, (2010).

\bibitem{Jiao2006}
Jiao, F., Wang, S., Lee, C-h, Greiner, R. and Schuurmans, D. 'Semi-supervised conditional random fields for improved sequence segmentation and labeling', in {\em ICCL2006} (2006).

\bibitem{Kalnay}
Kalnay et al., \newblock The NCEP/NCAR 40-year reanalysis project, {\em Bull. Amer. Meteor. Soc.},
77, (1996)

\bibitem{Lafferty_2001}
Lafferty, J.D, McCallum, A. and Pereira, F. C. N., Conditional Random Fields: Probabilistic Models for Segmenting and Labeling Sequence Data. In Proc. of ICML01,(2001).

\bibitem{Lee2012MI}
Lee, K. and Carlin, J. 'Recovery of information from multiple imputation: a simulation study', {\em Emerging Themes in Epidemiology}, (9) , (2012)

%\bibitem{Mann2007a}
%Mann, G~S and McCallum, A. 'Efficient computation of entropy gradient for semi-supervised conditional random fields.', in {\em HLT-NAACL (Short Papers)}, pp. 109--112, (2007).

\bibitem{Mann2007b}
Mann, G~S and McCallum, A., 'Simple, robust, scalable semi-supervised  learning via expectation regularization', in {\em Proc. of the 24th ICML}, pp. 593--600, (2007).

\bibitem{Mcknight2007}
Mcknight, P~E., Mcknight, K~M., Sidani, S. and Figueredo, A~J. {\em Missing Data: A Gentle Introduction (Methodology In The Social Sciences)}, {The Guilford Press}, (2007).

\bibitem{Menne09}
Menne, M~J., Williams, C. N.~Jr. and Vose, R~S. The US historical climatology network monthly temperature data (2009)

\bibitem{Mohan2013}
Mohan, K., Pearl, J. and Tian, J., 'Graphical models for inference with missing data', in {\em NIPS 26} 1277--1285, (2013).

\bibitem{Nastos13}
Nastos, P.T., Moustris, K.P., Larissi, I.K., Paliatsos, A.G., 'Rain intensity forecast using Artiﬁcial Neural Networks in Athens Greece', {\em Atmospheric Research 119}, 153-160, (2013).

\bibitem{Ng01}
Ng, A.Y., Jordan, M.I., 'On Discriminative vs. Generative Classifiers: A comparison of logistic regression and naive Bayes', {\em NIPS 2001}, MIT Press, pp. 841--848, (2001).

\bibitem{Ouzienko2011}
Ouzienko, V. and Obradovic, Z., 'Imputation of missing links and attributes in longitudinal social surveys', {\em Machine Learning}, 95(3), pp. 329--356, Springer US,(2014).

\bibitem{Qin2008}
Qin, T., Liu, T.-Y., Zhang, X.-D., Wang, S., and Li, H. 'Global ranking using continuous conditional random fields.', in {\em NIPS}, pp. 1281--1288, (2008).

\bibitem{Radosavljevic2010}
Radosavljevic, V., Vucetic, S.  and Obradovic, Z. 'Continuous conditional random fields for regression in remote sensing', in {\em ECAI}, pp. 809--814, (2010).

\bibitem{Rubin87}
Rubin D.B., 'Multiple Imputation for Nonresponse in Surveys (Wiley Series in Probability and Statistics)', (1987).

\bibitem{Salakhutdinov2003}
Salakhutdinov, R., Roweis, S.T., Ghahramani, Z., 'Optimization with EM and expectation-conjugate-gradient', in {\em ICML 2003}, pp. 672, (2003).

\bibitem{Silverman00}
Silverman, D., Dracup, J.A., 'Artificial Neural Networks and Long-Range Precipitation Prediction in California', {\em J. Appl. Meteor. 39}, 57-66, (2000).

\bibitem{Sokolovska2011}
Sokolovska, N. 'Aspects of semi-supervised and active learning in conditional random fields.', in {\em ECML/PKDD (3)}, volume 6913, pp. 273--288. Springer, (2011).

\bibitem{Verbeek}
Verbeek, J.J. and Vlassis, N., Gaussian fields for semi-supervised regression and correspondence learning. Pattern Recogn. 39, 10, 1864--1875, (2006).

\bibitem{Williams05}
Williams D., Liao X., Xue Y., Carin L., 'Incomplete-data classification using logistic regression'. {\em ICML2005 (2005)}.

\bibitem{Zhu03}
Zhu, X., Ghahramani, Z. and Lafferty, J., 'Semi-supervised learning using gaussian fields and harmonic functions', in {\em IN ICML}, pp. 912--919, (2003).


%\bibitem{AAAI13}
%Ristovski K., Radosavljevic V., Vucetic S., Obradovic Z., 'Continuous Conditional Random Fields for Efficient Regression in Large Fully Connected Graphs,' {\em Proc. The Twenty-Seventh (AAAI-13)}, Bellevue, Washington, July (2013).


%\bibitem{Lee12}
%Lee K.J., Carlin J.B., 'Recovery of information from multiple imputation: a simulation study', Emerg Themes Epidemiol, 9(3), (2012).


\end{thebibliography}
\end{document}